\documentclass{article}

\usepackage{PRIMEarxiv}

\usepackage[utf8]{inputenc} 
\usepackage[T1]{fontenc}    
\usepackage{hyperref}       
\usepackage{url}            
\usepackage{booktabs}       
\usepackage{amsfonts}       
\usepackage{amsmath}        
\usepackage{nicefrac}       
\usepackage{microtype}      
\usepackage{lipsum}
\usepackage{fancyhdr}       
\usepackage{graphicx}       
\usepackage{float}          
\graphicspath{{media/}}     

\title{Interpretable Perturbation Modeling Through Biomedical Knowledge Graphs}

\author{
  Pascal Passigan, Kevin Zhu, Angelina Ning  \\
   Massachusetts Institute of Technology \\
  \texttt{\{ppxscal, kbzhu, angn\_731\}@mit.edu} \\
}

\begin{document}
\maketitle

\begin{abstract}
Understanding how small molecules perturb gene expression is essential for uncovering drug mechanisms, predicting off-target effects, and identifying repurposing opportunities. While prior deep learning frameworks have integrated multimodal embeddings into biomedical knowledge graphs (BKGs) and further improved these representations through graph neural network message-passing paradigms, these models have been applied to tasks such as link prediction and binary drug-disease association, rather than the task of gene perturbation, which may unveil more about mechanistic transcriptomic effects. To address this gap, we construct a merged biomedical graph that integrates (i) PrimeKG++, an augmentation of PrimeKG containing semantically rich embeddings for nodes with (ii) LINCS L1000 drug and cell line nodes, initialized with multimodal embeddings from foundation models such as MolFormerXL and BioBERT. Using this heterogeneous graph, we train a graph attention network (GAT) with a downstream prediction head that learns the delta expression profile of over 978 landmark genes for a given drug-cell pair. Our results show that our framework outperforms MLP baselines for differentially expressed genes (DEG) --- which predict the delta expression given a concatenated embedding of drug features, target features, and baseline cell expression --- under the scaffold and random splits. Ablation experiments with edge shuffling and node feature randomization further demonstrate that the edges provided by biomedical KGs enhance perturbation-level prediction. More broadly, our framework provides a path toward mechanistic drug modeling: moving beyond binary drug-disease association tasks to granular transcriptional effects of therapeutic intervention.

\end{abstract}

\section{Introduction}

In the field of AI-driven drug discovery, drug repurposing is a critical area of research because it offers a cheaper, lower-risk pathway to identifying effective therapeutics for diseases. Binary prediction of drug-disease associations has long dominated the field, but this approach provides no information about off-target effects or mechanism-of-action (MOA). As such, this has given rise to phenotype-based methods, in which transcriptomics data play a critical role. This includes databases such as the Library of Integrated Network-based Cell-Signature (LINCS), which contains gene expression profiles for 42,080 perturbations across diverse cell lines. Although these databases exist, the vast number of possible drug-cell line interactions creates a combinatorial challenge that makes exhaustive experimental testing impractical. Therefore, computational models that can learn from existing data and generalize to new interactions are essential.

Recent approaches to drug repurposing have increasingly turned to graph-based and representation-learning methods to capture the complex relationships among drugs, targets, and diseases. Research by Dang and Ngyuen et al.  \cite{dang2025multimodal} leveraged biomedical knowledge graphs (BKGs) and pretrained biological language models to create richer representations of drugs, genes, diseases, and other biological entities. The researchers augmented PrimeKG, a biomedical knowledge graph (BKG) containing 17,080 diseases and 4,050,249 relationships, including protein-disease and drug-disease interactions, with multimodal embeddings that fused embeddings from BioBERT, ProteinBERT, DNABert, and MolFormer, to create PrimeKG++, which outperformed random initialization and direct LM-derived embeddings at downstream link prediction tasks. Zhao et al.\cite{zhao2025rgldr} leveraged GNN-based drug reasoning: they similarly initialized multi-modal node embeddings with pre-trained models, then learned representations from two graphs, one encoding drug-disease interactions and another encoding drug-similarity relations, to predict drug-disease associations. Their model outperformed state-of-the-art baselines on three benchmark drug-disease datasets: Cdataset, Gdataset, and LRSSL. These recent works point to the power of fusing multi-modal embeddings from pre-trained biological language models, using GNN message-passing to learn more contextualized embeddings, and leveraging the complex biological relations encoded in BKGs in learning drug-disease associations.

In the realm of gene perturbation prediction, modeling the combinatorial space of interactions between chemicals and cell lines is a complex challenge. Some existing supervised learning models include DLEPS, which predicts gene expression without distinguishing between cell types; DeepCE and CIGER, which use one-hot encoding to distinguish between cell types; and MultiDCP, which uses cellular context that enables predictions for novel cell lines. However, supervised learning models may struggle to distinguish between true signal and the inherent noise of gene expression data, which has led to variational autoencoder (VAE) based frameworks that denoise and reconstruct perturbation profiles. TranSiGen, developed by Tong et al.\cite{tong2024deep}, is one such model, which trains two VAEs in parallel using the LINCS 2020 dataset (one to learn baseline gene expressions and another to learn perturbed gene expressions) and learns a mapping from the baseline expression and drug compound representation to the perturbed expression. It outperformed existing perturbation predictors, such as the aforementioned DLEPS, DeepCE, CIGER, and MultiDCP and was used to identify drug repurposing hits for pancreatic cancer that were experimentally validated in vitro.

Although previous studies have leveraged individual components such as biomedical knowledge graphs like PrimeKG, multimodal embeddings from biological language models, graph neural network message-passing frameworks for drug-disease association, and LINCS L1000 transcriptomic profiles for perturbation modeling, no existing framework has yet integrated all of these elements to predict perturbation profiles.

\section{Methods}

Our framework integrates BKGs through the use of PrimeKG++; feature rich multimodal embeddings from BioBERT, ProtBERT, and MolFormer; graph attention network (GAT) message-passing to capture higher-order dependencies among biological entities; and LINCS 2020 transcriptomic data to predict drug-induced perturbations while elucidating mechanistic pathways through graph edges and gene expression profiles. We merged PrimeKG++, the compounds, and the corresponding metadata, investigated in the LINCS dataset into a unified heterogeneous directed graph to enable the model to query the graph's relations and features. This included producing SMILES strings, reconciling the additional drugs from LINCS with the existing ones from PrimeKG++, and producing edges between the drugs and the genes they were indicated to target. We aim to learn a function that takes a knowledge graph, a query drug in SMILES form, a baseline cell line expression profile, and predicts a delta vector to model a perturbation event.

Each node in our graph is represented by a [2, 768] embedding tensor, to capture two complementary modalities, with the exception of the cell line nodes which described the expression profile through the 978 landmark genes, as described by Table~\ref{tab:multimodal}. The final heterogeneous graph contains seven node types: drugs, genes/proteins, diseases, biological processes, molecular functions, cellular components, pathways, and cells. Edge types encode diverse biological relationships including drug-protein targeting, protein-protein interactions, GO term associations, disease-gene links, and drug-drug structural similarity.

\begin{table}[h]
\centering
\begin{tabular}{|p{3.5cm}|p{3.5cm}|p{3.5cm}|}
\hline
\multicolumn{3}{|c|}{\textbf{Multimodal Embeddings [2,  768]}} \\
\hline
\textbf{Drugs} \newline N=13,611 & 
\textbf{MolFormer-XL} \newline SMILES & 
\textbf{BioBERT v1.1} \newline Functional Description \\
\hline
\textbf{Gene/Protein} \newline N=21,597 & 
\textbf{ProtBERT} \newline Amino Acid seq & 
\textbf{BioBERT v1.1} \newline Functional Description \\
\hline
\textbf{Disease} \newline N=17,054 & 
\textbf{BioBERT v1.1} \newline Clinical Definition & 
\textbf{BioBERT v1.1} \newline Clinical Description \\
\hline
\textbf{GO Terms} \newline N=46,485 & 
\textbf{BioBERT v1.1} \newline Term Definition & 
\textbf{BioBERT v1.1} \newline Duplicated \\
\hline
\textbf{Cell Expression Vector} \newline [2,978] & 
\textbf{LINCS} \newline Baseline Expression & 
\textbf{LINCS} \newline Baseline Expression \\
\hline
\end{tabular}
\caption{Multimodal embeddings structure}
\label{tab:multimodal}
\end{table}

\subsection{Model Architecture}

All models in our framework share a common arithmetic structure inspired by ChemCPA~\cite{hetzel2022predicting}: rather than directly predicting gene expression, each model predicts a perturbation delta $\Delta$ that is added to the baseline expression. This formulation disentangles the drug effect from the cellular context:
\begin{equation}
\hat{\mathbf{y}} = \mathbf{b}_c + \Delta
\end{equation}
where $\mathbf{b}_c \in \mathbb{R}^{978}$ is the baseline expression profile for cell line $c$ and $\hat{\mathbf{y}} \in \mathbb{R}^{978}$ is the predicted perturbed expression.

\paragraph{MLP Baseline.} The first baseline uses only drug features and cell baseline, with no graph structure:
\begin{align}
\mathbf{h}_{\text{drug}} &= g_{\text{enc}}(\mathbf{x}_{\text{drug}}) \in \mathbb{R}^{256} \\
\Delta &= f_{\Delta}(\mathbf{h}_{\text{drug}} \oplus \mathbf{b}_c) \in \mathbb{R}^{978} \\
\hat{\mathbf{y}} &= \mathbf{b}_c + \Delta
\end{align}
where $\mathbf{x}_{\text{drug}} \in \mathbb{R}^{1536}$ is the flattened multimodal drug embedding (two 768-dimensional modalities concatenated), $g_{\text{enc}}: \mathbb{R}^{1536} \rightarrow \mathbb{R}^{256}$ is a 2-layer drug encoder with batch normalization and dropout, $\mathbf{b}_c \in \mathbb{R}^{978}$ is the baseline expression, and $\oplus$ denotes concatenation. The delta predictor $f_{\Delta}: \mathbb{R}^{1234} \rightarrow \mathbb{R}^{978}$ is a 3-layer MLP.

\paragraph{MLP+Targets Baseline.} The second baseline additionally incorporates mean-pooled target protein embeddings:
\begin{align}
\mathbf{h}_{\text{drug}} &= g_{\text{enc}}(\mathbf{x}_{\text{drug}}) \in \mathbb{R}^{256} \\
\mathbf{h}_{\text{targets}} &= g_{\text{enc}}(\mathbf{x}_{\text{targets}}) \in \mathbb{R}^{256} \\
\Delta &= f_{\Delta}(\mathbf{h}_{\text{drug}} \oplus \mathbf{h}_{\text{targets}} \oplus \mathbf{b}_c) \in \mathbb{R}^{978} \\
\hat{\mathbf{y}} &= \mathbf{b}_c + \Delta
\end{align}
where $\mathbf{x}_{\text{targets}} = \frac{1}{|T|} \sum_{t \in T} \mathbf{x}_t \in \mathbb{R}^{1536}$ is the mean of embeddings for drug target proteins $T$. The concatenated input to $f_{\Delta}$ is 1490-dimensional ($256 + 256 + 978$).

\paragraph{Graph Attention Network (GAT).} Our main model employs GATv2 \cite{brody2022attentive} for heterogeneous message passing:
\begin{align}
\mathbf{h}^{(0)} &= g_{\text{enc}}(\mathbf{x}) \quad \text{(per node type, } \mathbb{R}^{1536} \rightarrow \mathbb{R}^{256}\text{)} \\
\mathbf{z}_{\text{drug}} &= \text{GATv2}(\mathbf{h}^{(0)}, \mathcal{E}) \in \mathbb{R}^{256} \\
\Delta &= f_{\Delta}(\mathbf{z}_{\text{drug}} \oplus \mathbf{b}_c) \in \mathbb{R}^{978} \\
\hat{\mathbf{y}} &= \mathbf{b}_c + \Delta
\end{align}
where $\mathbf{z}_{\text{drug}}$ is the drug's graph-contextualized embedding after message passing, and $\mathcal{E}$ denotes the edge set. Feature encoders $g_{\text{enc}}$ are 3-layer MLPs ($1536 \rightarrow 1024 \rightarrow 1024 \rightarrow 256$) specific to each node type. GATv2 uses 4 attention heads with separate learned attention per edge type.

\subsection{Experimental Setup}

For data splitting, we employ scaffold-based splitting using Bemis-Murcko scaffolds to ensure rigor in generalization testing. All molecules sharing the same core scaffold are assigned exclusively to either train or test, preventing the model from exploiting structural similarities between splits. This is substantially harder than random splitting but provides realistic estimates of performance on structurally novel compounds. We use an 80/20 train/test split with a fixed random seed for reproducibility. In training, we used AdamW optimization with a learning rate of 0.001, batch size 512, and MSE loss on the 978 predicted gene expression values. Mini-batches are constructed via neighbor sampling (20 and 10 neighbors per hop) to enable scalable training. We train for 20 epochs, checkpointing the model with the best DEG correlation on the test set. To better understand if the graph structure helps with this task, we additionally set up ablation experiments, shuffling the edges and randomizing the node features to isolate the contributors of the signal, and comparing scaffold split to a randomized split.

\subsection{Evaluation}

We report two correlation-based metrics computed per sample and averaged: (1) Global Pearson correlation between predicted and observed expression across all 978 genes, and (2) DEG correlation computed only on the top 50 genes by absolute expression magnitude, focusing evaluation on the most strongly perturbed (biologically meaningful) genes rather than unchanged background. To contextualize model performance, we assess whether model differences are statistically significant using paired bootstrap testing over 1000 iterations. For each comparison we resample test samples with replacement and compute the difference in mean correlation, yielding 95\% confidence intervals and one-sided p-values for a granular quantitative comparison.

\section{Results}

\subsection{Main Results}

\begin{figure}[H]
\centering
\includegraphics[width=0.95\textwidth]{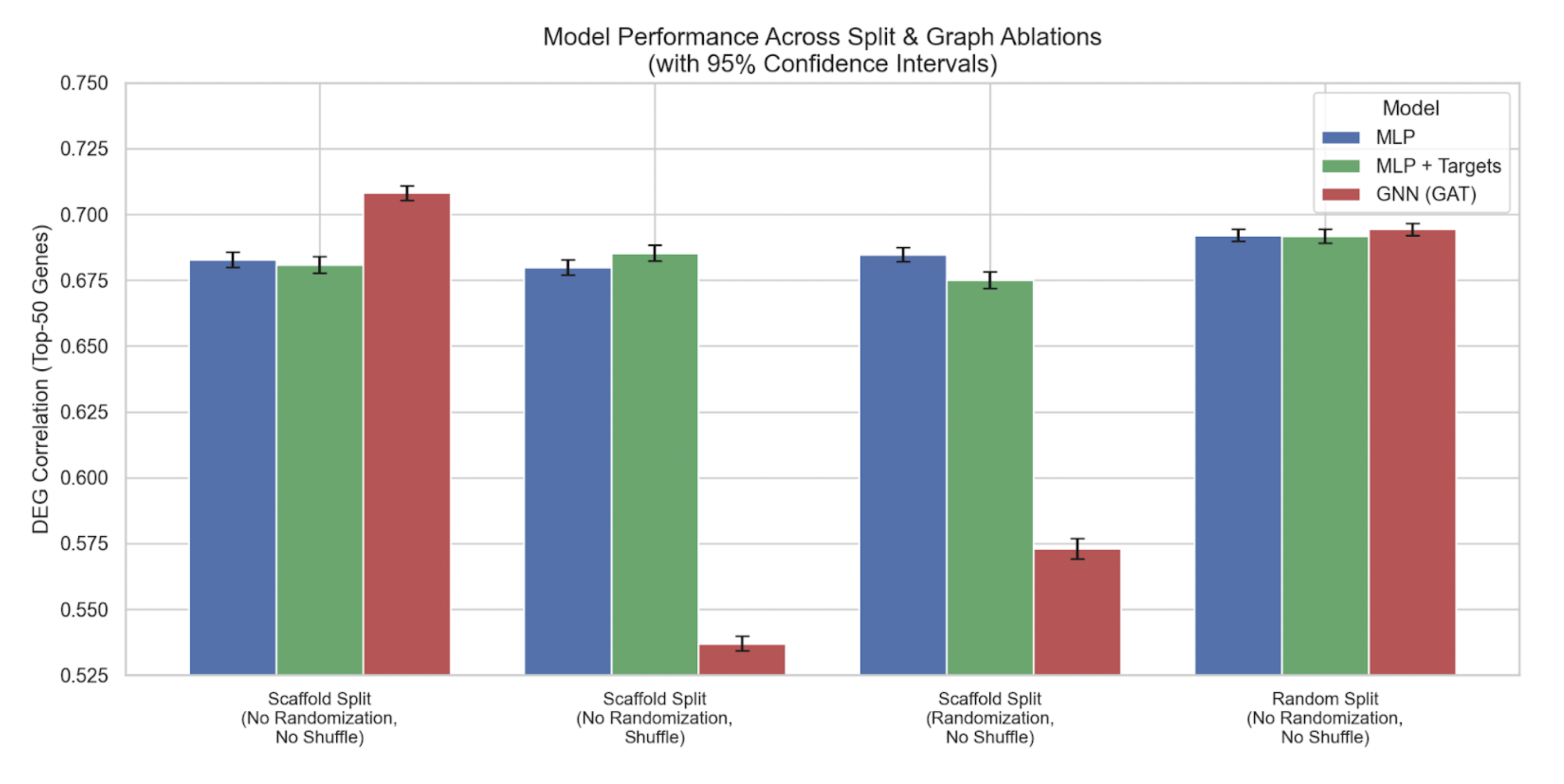}
\caption{Model performance across split types and graph ablations. Under scaffold split (leftmost cluster), GAT (red) achieves DEG correlation of 0.708, significantly outperforming MLP baselines ($\sim$0.68). Edge shuffling and node randomization (middle clusters) cause GAT performance to collapse below baseline levels, demonstrating the critical role of both graph topology and pretrained features. Under random split (rightmost), all models converge to similar performance ($\sim$0.69), indicating that graph structure provides advantages only when generalizing to novel chemical scaffolds.}
\label{fig:performance}
\end{figure}

Under the Bemis-Murcko scaffold split, the GAT model substantially outperformed both MLP baselines. As shown in Figure~\ref{fig:performance}, the GAT achieved a DEG correlation of 0.708 $\pm$ 0.001 (95\% CI: 0.706--0.711), compared with 0.683 $\pm$ 0.002 for the MLP and 0.681 $\pm$ 0.002 for the MLP+Targets model. Bootstrap tests confirmed all GAT--baseline differences as statistically significant ($p < 0.001$). This demonstrates that incorporating structured biological context improves generalization to unseen chemical scaffolds.

\subsection{Ablation Studies}

To better understand the contribution of graph structure and multimodal node features, we conducted a series of ablation experiments.

\subsubsection{Edge Shuffle}

When graph edges were shuffled---removing all meaningful biological relationships while preserving node features---GAT performance dropped from 0.708 to 0.537, falling below even the MLP baselines. This confirms that the relational structure encoded in the biomedical knowledge graph is essential for the GAT's improved generalization.

\subsubsection{Node Feature Randomization}

Randomizing all node embeddings except for the drug and cell representations produced a similar performance collapse (DEG correlation: 0.572), despite leaving the graph topology intact. Without semantically meaningful representations of genes, proteins, pathways, and diseases, the model was unable to exploit graph connectivity for mechanistic reasoning. The parallel degradation across this ablation and the edge shuffle highlights the complementary importance of both topology and pretrained multimodal features.

\subsubsection{Random Molecular Split}

Under a random split, all models performed nearly identically (MLPs $\approx$ 0.69; GAT = 0.694). Because test compounds in this setting often share scaffolds with the training set, models can rely on chemical similarity rather than leveraging graph structure. This result demonstrates that scaffold splitting is necessary to assess a model's ability to generalize to novel chemical space, and explains why the benefits of graph reasoning only emerge in the scaffold-split evaluation.

\subsection{Attention Analysis}

\begin{figure}[H]
\centering
\includegraphics[width=0.85\textwidth]{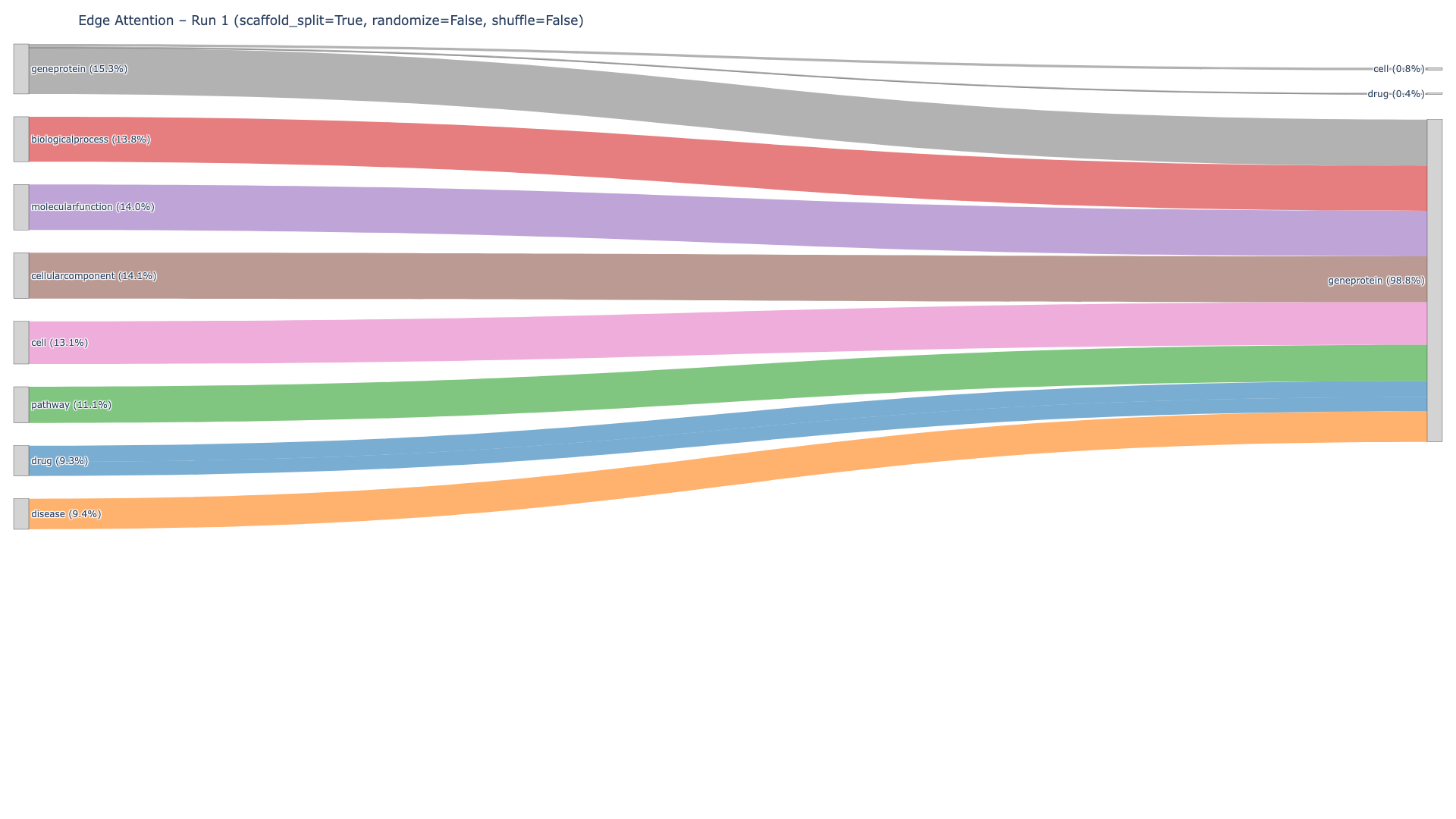}
\caption{\textbf{Baseline attention distribution (scaffold split, no ablation).} In the full-information setting, 98.8\% of attention routes through protein nodes, mirroring the biological reality that drugs act primarily via protein targets. This protein-centric pattern emerges without explicit supervision, demonstrating that the model learns mechanistically coherent reasoning from graph structure alone.}
\label{fig:attention_baseline}
\end{figure}

\begin{figure}[H]
\centering
\includegraphics[width=0.85\textwidth]{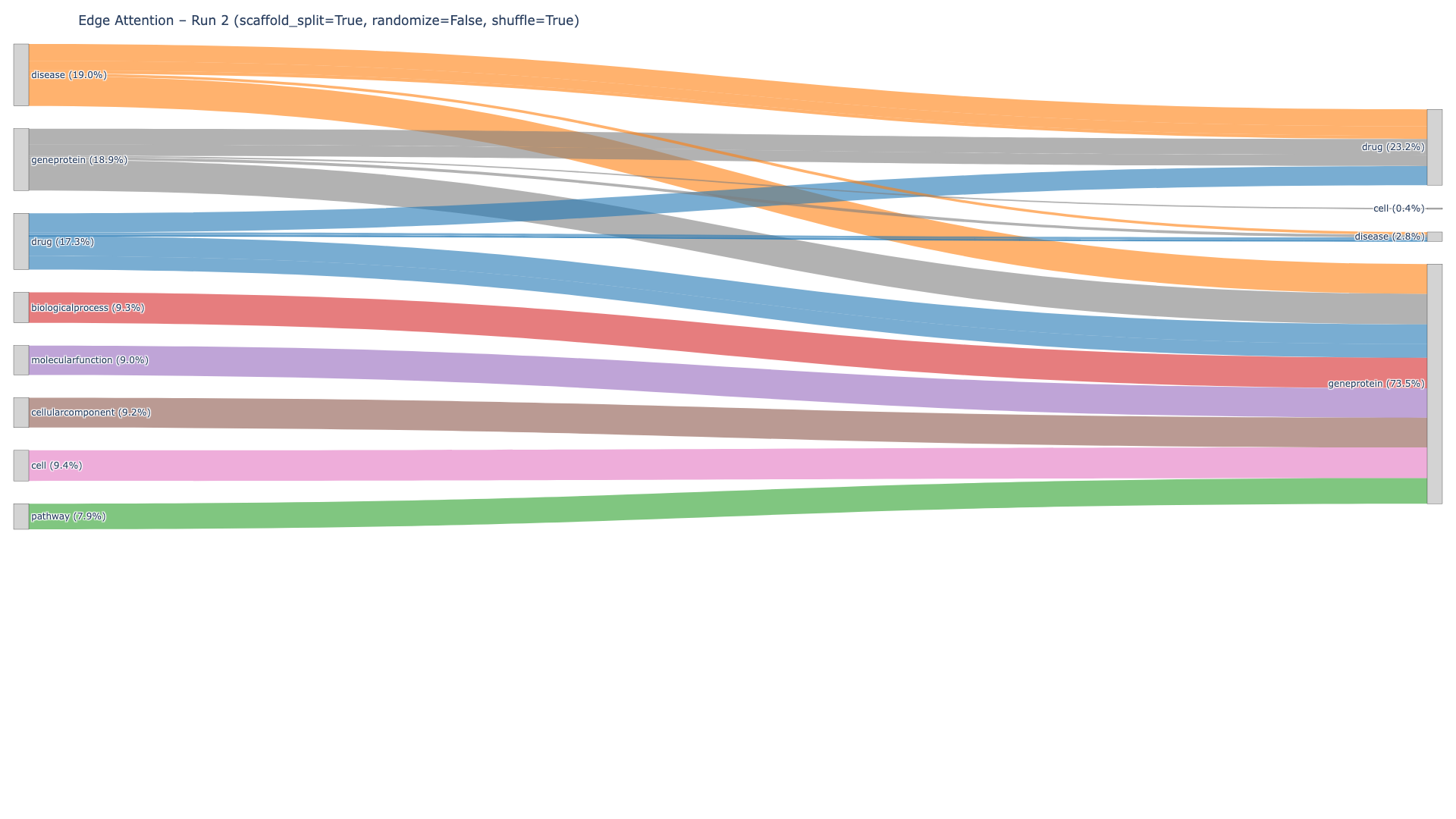}
\caption{\textbf{Attention distribution under edge shuffling.} Destroying biological relationships while preserving connectivity causes attention to fragment, with 23.2\% now flowing directly between drug nodes rather than through mechanistic intermediaries. This structural incoherence correlates with the performance collapse to 0.537 DEG correlation.}
\label{fig:attention_shuffle}
\end{figure}

\begin{figure}[H]
\centering
\includegraphics[width=0.85\textwidth]{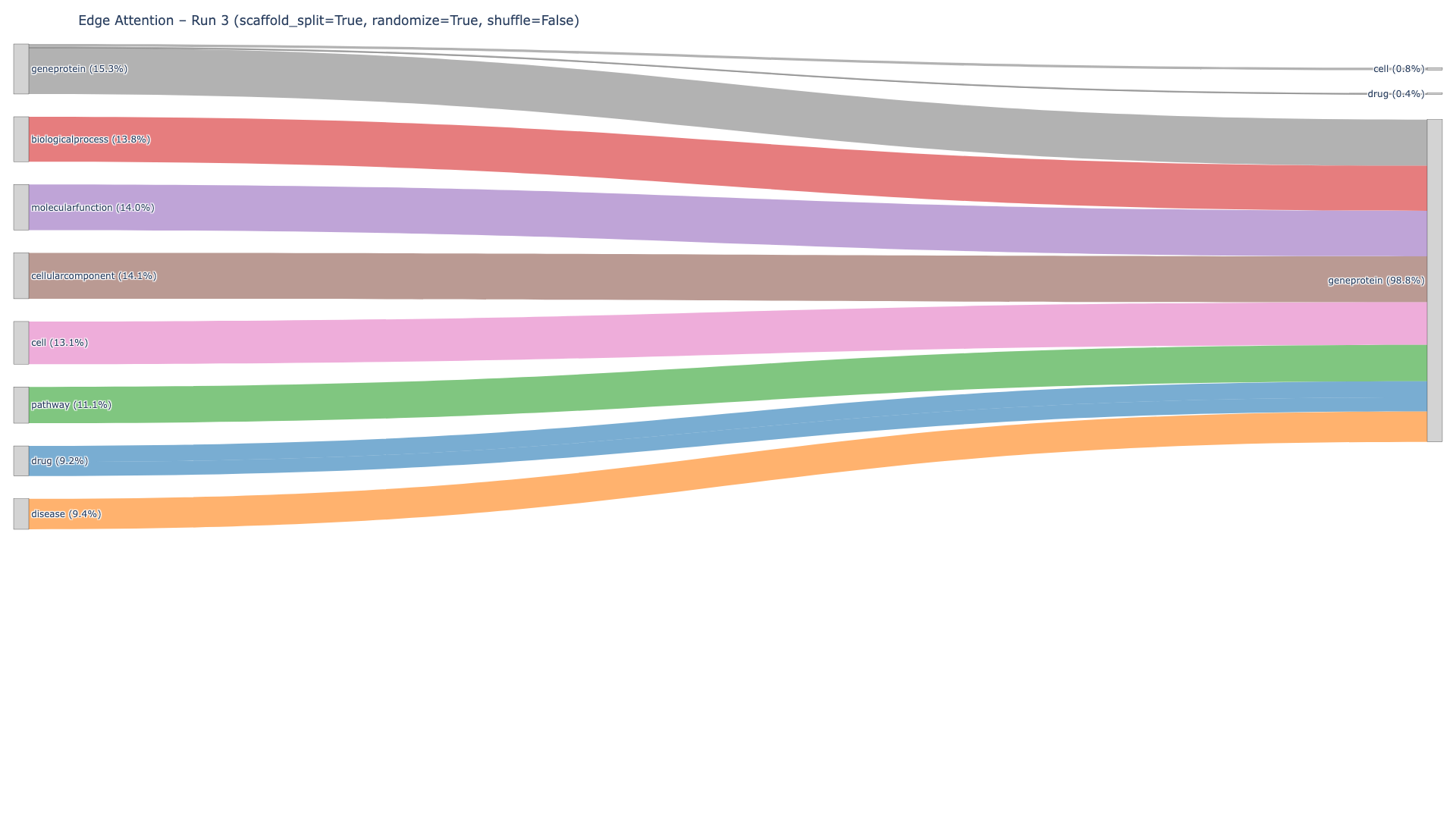}
\caption{\textbf{Attention distribution under node feature randomization.} Topology preserves the protein-centric attention flow, but without valid semantic embeddings, performance falls to 0.572 DEG correlation. This demonstrates that both graph structure \emph{and} meaningful node features are required for effective reasoning.}
\label{fig:attention_randomize}
\end{figure}

\begin{figure}[H]
\centering
\includegraphics[width=0.85\textwidth]{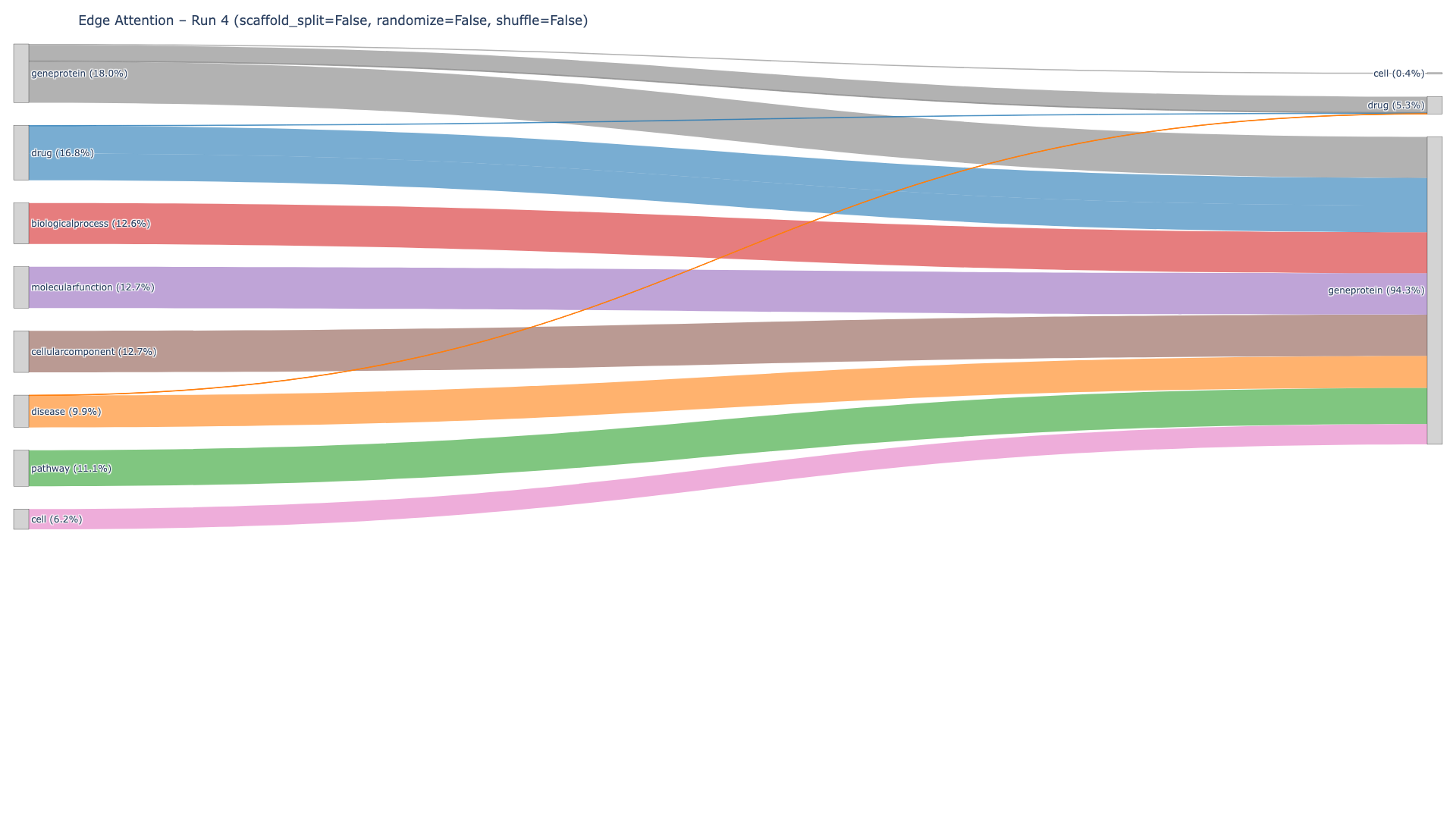}
\caption{\textbf{Attention distribution under random dataset split.} The model relies more heavily on drug-to-drug similarity when test compounds are chemically similar to training data. This explains why graph structure provides no advantage when chemical memorization suffices---the model bypasses mechanistic reasoning in favor of similarity matching.}
\label{fig:attention_random_split}
\end{figure}

To understand \emph{how} the model exploits structured biological knowledge, we analyzed attention distributions across four experimental conditions (Figures~\ref{fig:attention_baseline}--\ref{fig:attention_random_split}).

Together, these analyses validate that the GAT achieves its performance gains through learned biological reasoning chains that require both meaningful edges and pretrained embeddings.

\subsection{Case Study: Drug-Specific Reasoning Chains}

\begin{figure}[H]
\centering
\includegraphics[width=0.85\textwidth]{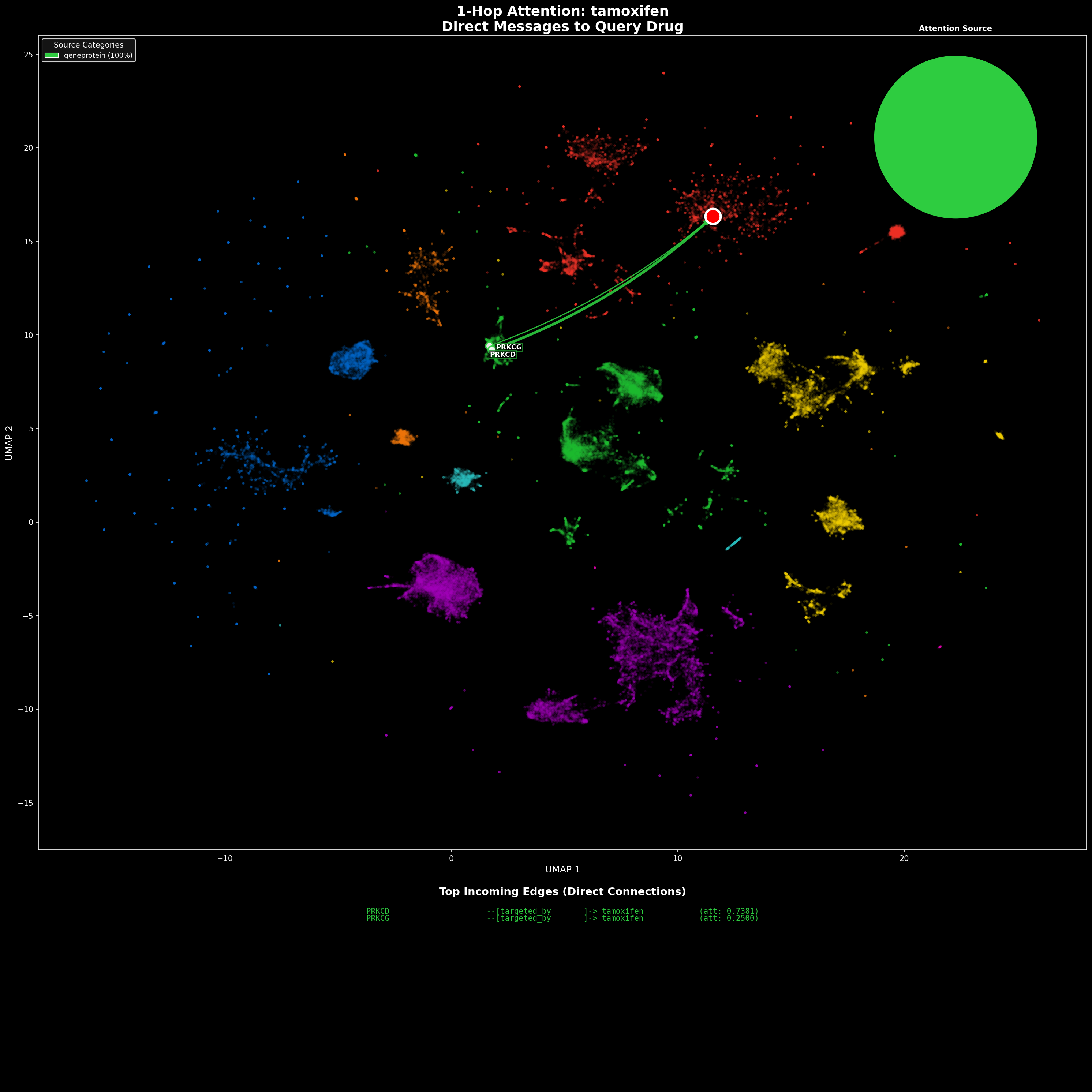}
\caption{\textbf{Direct attention for tamoxifen (1-hop).} The model allocates 100\% of direct attention to gene/protein nodes (green), consistent with tamoxifen's known mechanism of action through estrogen receptors (PRKCD, PRKCG visible in edge list). The UMAP projection clusters genes by biological function, with the query drug (white arrow) directly connecting to its protein targets. The attention source pie chart (top right) confirms exclusive protein-mediated reasoning at the first hop.}
\label{fig:tamoxifen_1hop}
\end{figure}

\begin{figure}[H]
\centering
\includegraphics[width=0.85\textwidth]{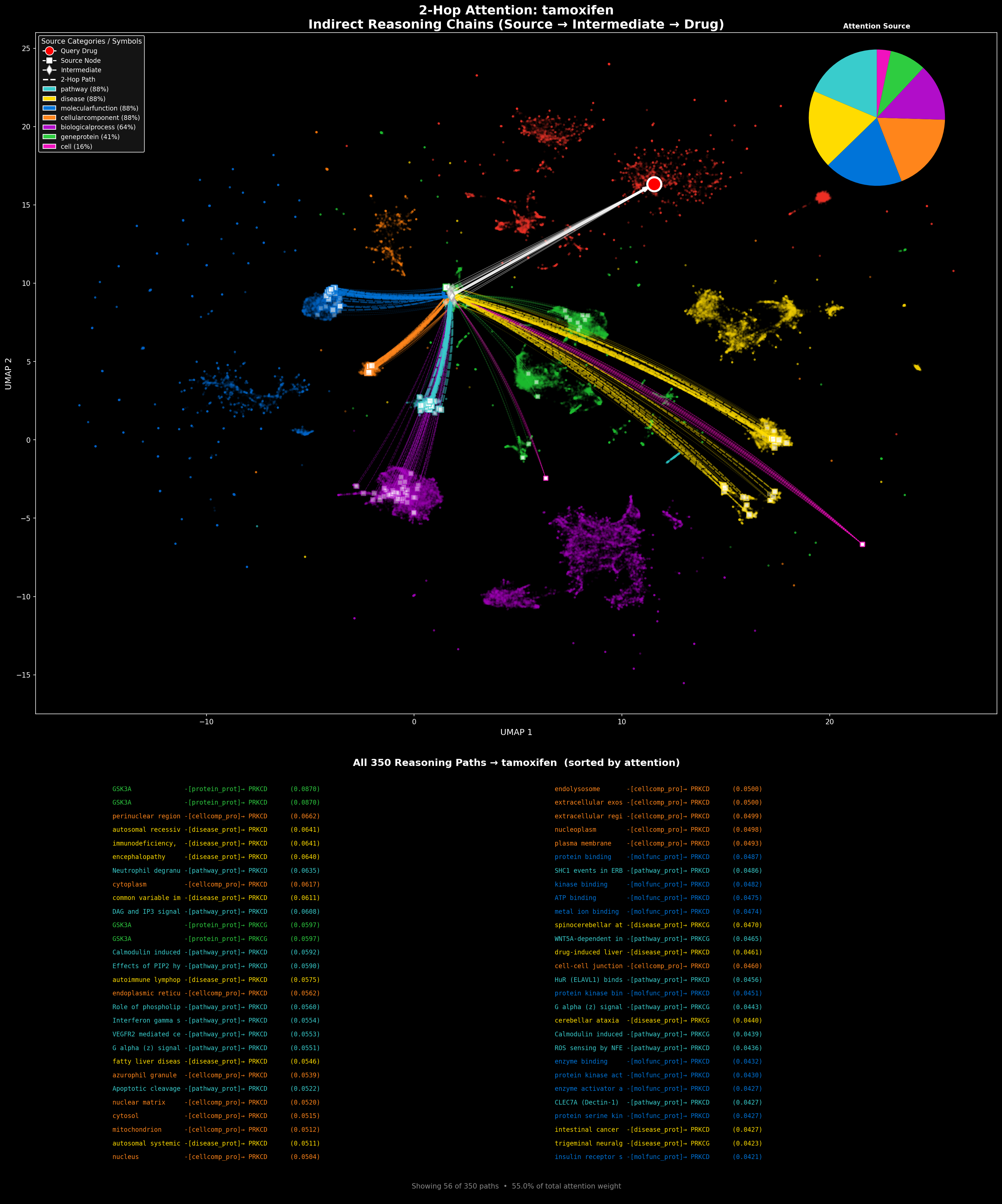}
\caption{\textbf{Indirect reasoning chains for tamoxifen (2-hop).} Extending to second-order neighbors reveals how the model propagates information through biological pathways. The attention now incorporates cellular components, biological processes, diseases, and molecular functions alongside proteins. The edge list shows specific mechanistic pathways. This multi-hop reasoning captures the drug's downstream transcriptional effects through interpretable biological intermediaries.}
\label{fig:tamoxifen_2hop}
\end{figure}

To illustrate how the model constructs mechanistic reasoning for individual drugs, we visualized attention-weighted graph neighborhoods for tamoxifen, a selective estrogen receptor modulator used in breast cancer treatment. Figure~\ref{fig:tamoxifen_1hop} shows direct (1-hop) connections, while Figure~\ref{fig:tamoxifen_2hop} reveals multi-step reasoning chains through intermediate biological entities.

These drug-specific visualizations demonstrate that our framework provides prediction-level interpretability: for any given drug-cell query, we can trace the exact reasoning path from molecular structure through protein targets and biological processes to predicted expression changes. This transparency is critical for validating model predictions and generating hypotheses about drug mechanisms.

\section{Discussion}

Our findings show that structured biomedical knowledge substantially improves perturbation prediction, but only when the evaluation setting demands generalization beyond the training distribution. Under random molecular splits, where many test compounds share scaffolds with training examples, the MLP baseline performed competitively. This behavior aligns with prior observations in drug response modeling: repeated exposure to chemically similar compounds allows a purely embedding-based model to succeed by interpolating within familiar chemical space rather than engaging with mechanistic biological context.

The scaffold split, by contrast, forces extrapolation to structurally novel compounds, and under this more realistic challenge the limitations of the MLP become apparent. Here, the GAT clearly outperformed both baselines, demonstrating that message passing across a merged biomedical knowledge graph provides complementary information not captured by drug embeddings alone. By integrating signals from genes, pathways, GO terms, diseases, and chemically related compounds, the GAT constructs richer representations that mitigate the sparsity and noise inherent in perturbation datasets. In this setting, relational structure contributes meaningful inductive bias, enabling the model to infer plausible biological mechanisms for unseen chemical structures.

The ablation studies reinforce this interpretation. When graph edges were shuffled, destroying biologically meaningful structure while preserving node features, GAT performance dropped sharply below MLP levels. Similarly, randomizing node embeddings produced a parallel collapse. These results demonstrate that neither topology nor multimodal features alone are sufficient; the predictive gains arise from their interaction. Furthermore, attention analyses illustrate this mechanism: in the full-information model, attention concentrates on protein and ontology nodes in patterns that mirror biological reasoning, whereas the shuffled-edge model exhibits incoherent attention.

Despite these advances, some limitations remain. First, although the GAT improves generalization under scaffold splits, the effect size is modest given the difficulty of predicting high-dimensional transcriptional responses. A key reason is the incompleteness and coarse granularity of existing biomedical knowledge graphs. Many edges encode broad associations but lack the context (cell type, condition, directionality) required for precise mechanistic reasoning. Our attention analyses also show that only a subset of edge types---primarily protein-protein interactions and gene ontology terms---meaningfully contribute to message passing, suggesting that large portions of the graph are either uninformative or not encoded in a way the model can leverage. Refining the graph to emphasize mechanistically relevant relations, or learning edge-specific importance weights, could substantially strengthen representation quality.

Second, our architectural exploration was limited to GATv2. Recent heterogeneous graph transformers, relation-aware message-passing models, and hybrid neural-symbolic architectures may more effectively exploit typed multimodal graphs and disentangle the relative contribution of different biological relations.

Beyond the specific results reported here, our framework establishes a generalizable paradigm for multimodal biomedical reasoning. The architecture is modular: foundation models serve as interchangeable feature extractors (MolFormer for molecules, BioBERT for text, ProtBERT for proteins), while the node-type-specific encoders project these heterogeneous embeddings into a shared representation space. The heterogeneous GATv2 layer learns relation-specific attention without manual feature engineering, allowing seamless incorporation of new edge types as biomedical ontologies evolve. This design enables straightforward extension to additional modalities—such as 3D protein structures, single-cell transcriptomics, or clinical phenotypes—by simply adding new foundation model embeddings and corresponding node types, without modifying the core message-passing architecture.

Finally, an intriguing future direction is the inverse perturbation problem: predicting candidate drugs capable of inducing a desired transcriptional effect. Because our graph encodes how drugs propagate information to genes, inverting this mapping (desired perturbation $\rightarrow$ drug) may be feasible with appropriate constraints and could support mechanism-guided drug design or repurposing workflows.

\section{Conclusion}

This work demonstrates that incorporating structured biological knowledge into drug perturbation models enables meaningful gains in generalization, particularly when evaluating on chemically novel compounds. Graph message passing enriches drug representations with contextual information from interacting genes and biological processes, producing improvements that neither unimodal embeddings nor graph structure alone can provide. At the same time, the modest effect sizes observed highlight both the complexity of transcriptomic prediction and the current limitations of available knowledge resources.

Future progress will require expanding and refining multimodal node representations, incorporating dynamic or perturbation-aware cellular context, and leveraging more expressive heterogeneous GNN architectures. Continued advances in biomedical knowledge graph construction---through automated literature extraction, causal modeling, and integration of single-cell perturbation data---may further enhance the capacity of graph-based models to capture mechanistic drug effects. Ultimately, our results underscore that structured biological knowledge is not merely an auxiliary signal but a necessary ingredient for building perturbation predictors capable of robust, mechanism-aware extrapolation across chemical space.

\section*{Code Availability}

Code and trained model weights will be made publicly available in future versions. 

\bibliographystyle{unsrt}  
\bibliography{references}

\end{document}